# Enhanced Edge-Perceptual Guided Image Filtering

Jinyu Li

*Abstract*—**Due to the powerful edge-preserving ability and low computational complexity, Guided image filter (GIF) and its improved versions has been widely applied in computer vision and image processing. However, all of them are suffered halo artifacts to some degree, as the regularization parameter increase. In the case of inconsistent structure of guidance image and input image, edge-preserving ability degradation will also happen. In this paper, a novel guided image filter is proposed by integrating an explicit first-order edge-protect constraint and an explicit residual constraint which will improve the edge-preserving ability in both cases. To illustrate the efficiency of the proposed filter, the performances are shown in some typical applications, which are single image detail enhancement, multi-scale exposure fusion, hyper spectral images classification. Both theoretical analysis and experimental results prove that the powerful edge-preserving ability of the proposed filter.**

## I. INTRODUCTION

The edge-preserving smoothing techniques are widely applied in image processing, computer vision and computational photography, such as tone mapping algorithms for high dynamic range image [1],image denoising [2], multi-scale exposure image fusion [3][4], structure extraction from texture [5], single image dehazing [6][7], etc. As its name suggests, the property of this type of technique is to smooth the image while preserving the edges. In other words, it has a good ability to distinguish the noise and realistic edge information of the input image. A series of edge-preserving smoothing filters have been proposed in recent decades. All of them can be broadly divided into two types: One is based on global optimization, such as total variation (TV) [8], weighted least squares based methods [9]-[11], and L0-norm gradient minimization [12]. The above algorithms are based on solving an optimization problem, which is the combination of real data term and regularization penalty term. By rationally designing and optimizing the structure of the equation, better results can often be obtained. However, all of these solutions are solved through multiple iterations and require reference to the global information of the image, so this kind of algorithms are often time-consuming. The other is based on local linear models. such as bilateral filter (BLF) [13], and part of its improved version[14][15], guided image filtering (GIF) [16], and part of its improved version [17][18]. They have an advantage in calculation time, but they tend to produce annoying halos at the edges in some specific scenarios. How to reduce this negative impact has been a research hotspot in recent years.

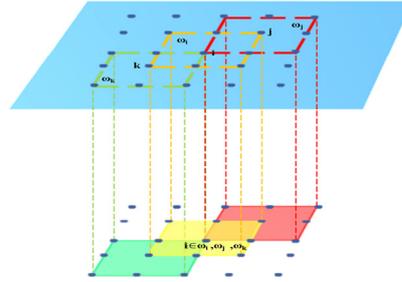

Fig. 1. Schematic illustration of the overlapping windows that cover pixel i.

Among the many proposed methods mentioned above, the most commonly used may be bilateral filter (BLF) [13] and guided image filter (GIF) [16] in image processing. Bilateral filtering (BLF), its structure resembles a combination of two Gaussian filter, updating value of the central pixel of a local window in the spatial domain which corresponds to pixel locations and in the intensity domain which corresponds to color or brightness values, respectively. The updated rule is that pixels which within preset neighborhoods of the central pixel are assigned Gaussian statistical values based on their correlation to the position and intensity of the central pixel. But because of the over-enhancing at the edge, it may suffer from gradient reversal artifacts near some edges when used for detail enhancement [1]. Thus, in order to avoid gradient reversal artifacts, Guided image filter (GIF) was proposed by applying a local linear model. Its innovation is that a guidance image is added to the calculation of the pixel value in the input image. Taking the advantage of the idea of integral graphs, its calculation time is greatly reduced compared to the BLF. However, since the regularization parameters are fixed, halos, which refer to areas of lighter or darker pixels that appear around edges, are prone to produce at some certain edges [16].

In order to improve the disadvantage that GIF is easy to produce halos at the edges, a series of improved algorithms have been proposed in recent years. These improvements can be divided into two main parts, one is to improve the edge-aware weighting, and the other is to improve the parameter averaging strategy. In order to improve the defect of regularization parameter fixation, in [17], a weighted guided image filter (WGIF) was proposed by introducing an edge-aware weighting to the regularization term and thus it performs better at the edges. Based on the research work of GIF and WGIF, in [18], gradient domain guided image filtering (GGIF) was proposed by adding an explicit first-order edge-aware constraint and a multi-scale edge-aware weighting to the local linear model, it preserves edges better than WGIF.

As shown in Fig.1, an arbitrary pixel i in the input image is covered by different overlapping windows. Hence the filtering output $R_i$ will get different values when it is computed in different windows. However, algorithms mentioned above all applied a simple averaging strategy to deal with the fact that each pixel is contained in several overlapping windows [16]. Since the same weights are added to the calculation of the final parameters $a_k$ and $b_k$, the difference between pixels is reduced, in other words, the edge-preserving ability of the mentioned filters is reduced. To solve this problem, a series of algorithms were proposed to reformulate the averaging step of GIF. In [21], by utilizing the steering kernel to learn local structure prior of the guidance image and setting it as the weighted average strategy to update the value of $a_k$ and $b_k$, a weighted guided image filter with steering kernel (SKWGIF) was proposed. In [22], a weighted aggregation for guided filter (WAGIF), which took mean square error (MSE) into the construction of weighted average strategy, can improve edge-preserving ability significantly.

In this paper, after an analysis of the GIF [16] and its improved versions mentioned above, to improve the edge-preserving ability, similar as the GGIF [18], an Enhanced Edge-Perceptual Guided Image Filter (EPGIF) is proposed by integrating an explicit first-order edge-protect constraint and an explicit residual constraint. Taking the local variance information and difference in brightness of the guidance image into consideration, a new edge-aware weighting is proposed in this paper for better performance. A new edge-protect constraint is constructed to constrain the value of $a_k$, so that $a_k$ is closer to 1 at the edges and closer to 0 in the flat areas. At the same time, in the average step, in order to make the value of $a_k$ closer to the real situation, an improved weighted average method is proposed. Similar to the GIF [16], WGIF [17], GGIF [18], the proposed filter also avoids gradient reversal effect and the complexity of the proposed filter is O(N) for an image with N pixels. Benefited from the low computational complexity, the proposed filter can be applied in many image processing scenarios, such as single image detail enhancement, image fusion, image classification. In the subjective visual evaluation and objective parameter evaluation, it is improved compared with the comparison algorithms. Generally speaking, the major contributions of this paper are: 1) a new edge-aware weighting; 2) a new edge-protect constraint; 3) an explicit residual constraint; 4) an improved weighted average strategy.

The reminder of this paper is organized as follows. In the next section, it is the related works about guided image filter. A gradient domain content adaptive guided image filter is proposed in section III. In section IV, the applications and the experiment results will be discussed. At the end, section V concludes this paper.

## II. RELATED WORKS

In the original version of GIF, let two images $G$ and $X$ be a guidance image and an input image to be filtered, respectively. They usually could be identical. In a square window $\zeta(p)$ centered at a pixel p of a radius $\zeta$, it is assumed that the output image R is a first-order linear transform of the guidance image G in the window $\zeta(p)$.

$$R(p) = a_{p'}G(p) + b_{p'}, \forall p \in \Omega_{\zeta}(p'), \quad (1)$$

where $a_{p'}$ and $b_{p'}$ are two coefficients according to the square window $\Omega_{\zeta}(p')$. In order to get the values of the two coefficients, the usual practice is to minimize a cost function $E(a_{p'}, b_{p'})$, which is defined as follows:

$$E = \sum_{p \in \Omega_{\zeta}(p')} \left[ \left( a_{p'}G(p) + b_{p'} - X(p) \right)^2 + \lambda a_{p'}^2 \right], (2)$$

where $\lambda$ is a regularization parameter to avoid $a_{p'}$ is too large. The optimal values of $a_{p'}$ and $b_{p'}$ are derived as the following (3) and (4), respectively:

$$a_{p'} = \frac{\mu_{G \odot X, \zeta}(p') - \mu_{G,\zeta}(p')\mu_{X,\zeta}(p')}{\sigma_{G,\zeta}^2(p') + \lambda}, \quad (3)$$

$$b_{p'} = \mu_{X,\zeta}(p') - a_{p'}\mu_{G,\zeta}(p'), \quad (4)$$

where $\odot$ represents the Hadamard product of two matrices. $\mu_{X,\zeta}(p), \mu_{G,\zeta}(p), \mu_{G \odot X,\zeta}(p)$ are the mean values of $X$, $G$, $G \odot X$ in the window $\Omega_{\zeta}(p')$, respectively.

However, due to the regularization parameter fixation, GIF has relatively limited ability to perceive edges, which directly leads to annoying halos at the edges. In WGIF, it performs better at the edges by introducing an edge-aware weighting $\Phi_G$ to the regularization term. The cost function in (2) is redefined as follows:

$$E = \sum_{p \in \Omega_{\zeta}(p')} \left[ \left( a_{p'}G(p) + b_{p'} - X(p) \right)^2 + \frac{\lambda}{\Phi_G(p')} a_{p'}^2 \right], (5)$$

where the edge-aware weighting $\Phi_G$ is defined by using local variances of 3×3 windows of all pixels as follows:

$$\Phi_G(p') = \frac{1}{N} \sum_{p=1}^{N} \frac{\sigma_{G,1}^2(p') + \varepsilon}{\sigma_{G,1}^2(p) + \varepsilon}, \quad (6)$$

Using this content adaptive function, the significance of image pixels at the edge is reflected well by the function value.

The optimal values of ap' and bp' are derived as the following (7) and (8), respectively:

$$a_{p'} = \frac{\mu_{G \odot X, \zeta}(p') - \mu_{G,\zeta}(p')\mu_{X,\zeta}(p')}{\sigma_{G,\zeta}^2(p') + \frac{\lambda}{\Phi_G(p')}}, \quad (7)$$

$$b_{p'} = \mu_{X,\zeta}(p') - a_{p'}\mu_{G,\zeta}(p'), \quad (8)$$

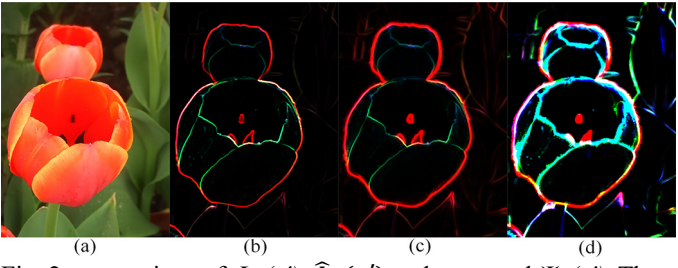

(a)      (b)      (c)      (d)

Fig. 2. comparison of $\Phi_G(p')$, $\hat{\Phi}_G(p')$ and proposed $\Psi_G(p')$. The window size is $33\times33$ for (c), (d), and the values are divided by 1000. (a) input image. (b) $\Phi_G(p')$. (c) $\hat{\Phi}_G(p')$. (d) $\Psi_G(p')$.

To further exempting edges from the effects of smoothing, an explicit constraint was introduced to strengthen the value of edge retention factor $a_{p'}$. The cost function in (5) is redefined as follows:

$$E = \sum_{p\in\Omega_\zeta(p')} \left[ \left(a_{p'}G(p)+b_{p'}-X(p)\right)^2 + \frac{\lambda}{\hat{\Phi}_G(p')}\left(a_{p'}-\gamma_{p'}\right)^2 \right], \quad (9)$$

where the edge-aware weighting $\hat{\Phi}_G$ is defined in (10) by using standard deviation of $3\times3$ windows and $(2\zeta+1) \times (2\zeta+1)$ windows of all pixels. And explicit constraint $\gamma_{p'}$ is defined in (11) by applying a sigmoid function, respectively computed as:

$$\hat{\Phi}_G(p') = \frac{1}{N}\sum_{p=1}^{N}\frac{\chi(p')+\varepsilon}{\chi(p)+\varepsilon}, \quad (10)$$

$$\gamma_{p'} = 1 - \frac{1}{1+e^{\frac{4(\chi(p')-\mu_{\chi,\infty})}{(\mu_{\chi,\infty}-\min(\chi(p')))}}}, \quad (11)$$

where $\chi(p')$ is defined as $\sigma_{G,1}(p')\sigma_{G,\zeta}(p')$, $\zeta$ is the window size of the proposed filter. $\mu_{\chi,\infty}$ is the mean value of all $\chi(p')$. Compare to the edge-aware weighting $\Phi_G$ in WGIF, the edge-aware weighting $\hat{\Phi}_G$ in GGIF referenced the variance information at multiple scales, which makes the edge estimation information more accurate.

The optimal values of $a_{p'}$ and $b_{p'}$ are derived as the following (12) and (13), respectively:

$$a_{p'} = \frac{\mu_{G\odot X,\zeta}(p')-\mu_{G,\zeta}(p')\mu_{X,\zeta}(p')+\frac{\lambda}{\hat{\Phi}_G(p')}\gamma_{p'}}{\sigma^2_{G,\zeta}(p')+\frac{\lambda}{\hat{\Phi}_G(p')}}, \quad (12)$$

$$b_{p'} = \mu_{X,\zeta}(p') - a_{p'}\mu_{G,\zeta}(p'), \quad (13)$$

It can be seen from (12) that if the guidance image is not identical with the input image, the covariance of the guidance image and input image is usually not equal to the variance of the guidance image at the edges, therefore $a_{p'}$ is not equal to 1, which means sharp edge cannot be preserved well in this case.

For the final value of $\bar{a}_p, \bar{b}_p$, GIF, WGIF and GGIF are the same average strategy, which can be computed as follows, respectively:

$$\bar{a}_p = \frac{1}{|\Omega_\zeta(p)|}\sum_{p'\in\Omega_\zeta(p)}a_{p'}; \bar{b}_p = \frac{1}{|\Omega_\zeta(p)|}\sum_{p'\in\Omega_\zeta(p)}b_{p'}; \quad (14)$$

## III. PROPOSED ALGORITHM

Inspired by these algorithms, an improved guided filter with gradient domain constraint and residual constraint is proposed in this section. Experiment results show that its edge-preserving ability is better than GIF, WGIF and GGIF.

### A. A New Edge-Aware Weighting

Both WGIF and GGIF are easy to produce halos at the edges with the increase of regularization parameter $\lambda$ [19]. In the equation (7), when the pixel at the edges, the value of $\lambda/\Phi_G$ should be very small and not affect the relationship of covariance and variance in the equation. However, with the increase of $\lambda$, the value of edge-preserving weighting $\Phi_G$ in WGIF and $\hat{\Phi}_G$ in GGIF are too small to keep the property mentioned above, in other words, the robustness of the edge-preserving weighting is insufficient. Thus, increasing the value of the edge-preserving weighting is a feasible approach.

A new edge-aware weighting $\Psi_G(p')$ is defined by using local variances, average standard deviation, and mean value of $3\times3$ windows and $(2\zeta+1) \times (2\zeta+1)$ windows of all pixels as follows:

$$\Psi_G(p') = \frac{1}{N}\sum_{p=1}^{N}\frac{\varphi(p')+\varepsilon}{\varphi(p)+\varepsilon}, \quad (15)$$

where $N$ is the total number of pixels in the image $I$, and $\varepsilon$ is a small positive constant and its value is selected as $(0.001\times L)^2$ while $L$ is the dynamic range of the input image. What's more, $\varphi(p')$ is defined as $\frac{\sigma^2_{G,1}(p')\;\sigma^2_{G,\zeta}(p')}{\bar{\sigma}_{G,1}(p')\;\bar{\sigma}_{G,\zeta}(p')}$, $\zeta$ is the window size of the proposed filter.

$\varphi(p')$ contains multi-scale local variance information of the image. When the pixel is at an edge, its local variance value increases significantly in both scales, which can better judge the edge. To enhance it, with the idea of coefficient of variation (C.V.), the average local variance representing global information can be introduced into the local variance information. In this way, the problem of excessive differences in the distribution range of local variance values at different scales can be eliminated.

As shown in Fig.2, compared to the edge-aware weighting of the WGIF and the GGIF, the proposed one is brighter and has more edge information. Brighter colors represent larger values, which will help improve the robustness of the edge-aware weighting, in other words, as the regularization parameter $\lambda$ increases, it can better suppress the production of halo artifacts at the edges. In addition, local mean value and local variance are already calculated in the original GIF algorithm, there will only be a slightly increase in calculation time.

### B. A New Edge-Protect Constraint

In GGIF, a sigmoid function $\gamma$ is constructed for edge-protect constraint, which is defined in (11). As can be seen from the (12), it can regulate the value of $a_{p'}$, which means it has a significant impact on edge-preserving ability of the filter. After a simple algebra calculation, results can be obtained as follow:

*Case 1:* when $\chi(p')$ is equal to $\mu_{\chi,\infty}$, then the value of $\gamma_{p'}$ is equal to 1/2. Let $\gamma_{mean}$ represents the value of $\gamma_{p'}$ in this case.

*Case 2:* when $\chi(p')$ is equal to $min(\chi(p'))$, then the value of $\gamma_{p'}$ is equal to 0.0180, not zero. Let $\gamma_{min}$ represents the value of $\gamma_{p'}$ in this case.

For the Case 1, it is a basic fact that the value of $\gamma_{p'}$ must be greater than $\gamma_{mean}$ and closer to $\gamma_{max}$ when the pixel is at an edge. In (7), the possible values range of the $\gamma_{p'}$ corresponding to the pixels at the edges and in the flat area accounts for half of the total value range, respectively. However, for each image, it is different that the proportion of edge information contained in the entire image information, which means that the possible values range of the $\gamma_{p'}$ corresponding to the pixels at the edges should be adaptive according to the image content.

For the Case 2, when the pixel is in the flat area, because the pixel values tend to be consistent, $\chi(p')$ is significantly smaller than $\mu_{\chi,\infty}$, then the value of $\gamma_{p'}$ is close to $\gamma_{min}$. However, since $\gamma_{p'}$ cannot be 0 in this case, and the value of $\hat{\Phi}_G$ is small, it can be seen from (12) that when the regularization parameter $\lambda$ is large, $\lambda/\Phi_G$ becomes the main linear component, and the value of $a_{p'}$ is basically determined by $\gamma_{p'}$, in other words, the variance and covariance in (12) did not work. At the same time, it once again confirms that the value of edge-aware weighting should be increased.

In order to solve the problems above, an optimized *tanh* curve is proposed as follows:

$$\tau_{p'} = 0.5\tanh\left(\frac{2(\alpha(p') - \mu_{\alpha,\infty})}{(\mu_{\alpha,\infty} - \min(\alpha(p')))}\right) + c, \quad (16)$$

with

$$\tau_{p'} = \begin{cases} 0, & if \quad \tau_{p'} \leq 0 \\ c + \frac{(1-c)(\tau_{p'} - c)}{0.5}, & if \quad \tau_{p'} \geq c \\ \tau_{p'}, & else \end{cases} \quad (17)$$

where $\alpha(p')$ is $\varphi(p')\rho(p')$, $\mu_{\alpha,\infty}$ is the mean value of all $\alpha(p')$. And $c$ is a positive constant, the recommended value is between 0.1 and 0.45 and the default value is 0.35. After performing a piecewise operation in (17), it is easy to know that the minimum value $\tau_{min}$ is equal to 0 and the maximum value $\tau_{max}$ is equal to 1. Therefore, values of the $\tau_{p'}$ are closer to 1 at the edges and closer to 0 in the flat area. In other words, $a_{p'}$ is closer to 1 at the edges and closer to 0 in the flat area. This will improve the edge-preserving and flat area smoothing capabilities for the proposed filter.

### C. A New Weighted Average Strategy

Inspired by the WAGIF [22], a weight $W_{p'}$ with explicit residual constraint can be construct as follows:

$$W_{p'} = \exp^{-\frac{1}{|\Omega_\zeta(p)|}\sum_{p'\in\Omega_\zeta(p)}\frac{(\eta_{p'}(a_{p'}G(p)+b_{p'}-X(p)))^2}{\beta}}, \quad (18)$$

where $\beta$ is a positive constant, it is set to 1/500 instead of 1/200 in WAGIF, $\eta_{p'}$ is equal to $(1-\tau)$, and $|\Omega_\zeta(p')|$ is the cardinality of the set $\Omega_\zeta(p')$.

After a simple derivation, the calculation format of weight $W_{p'}$ is as follows:

$$W_{p'} = \exp^{-\frac{\eta_{p'}^2\left(\sigma_{X,\zeta}^2(p') - a_{p'}^2\sigma_{G,\zeta}^2(p')\right) - 2a_{p'}(a_{p'} - \tau_{p'})\frac{\lambda}{\Psi_G(p')}\eta_{p'}}{\beta}}, \quad (19)$$

When G and X are identical, (19) can be simplified as follows:

$$W_{p'} = \exp^{-\frac{\eta_{p'}^2\left((1-a_{p'}^2)\sigma_{G,\zeta}^2(p')\right) - 2a_{p'}(a_{p'} - \tau_{p'})\frac{\lambda}{\Psi_G(p')}\eta_{p'}}{\beta}}, \quad (20)$$

### D. The proposed filter

Based on the algorithms proposed above, the cost function in (5) can be redefined as follows:

$$E = \sum_{p\in\Omega_\zeta(p')}\left[\left(\eta_{p'}(a_{p'}G(p) + b_{p'} - X(p))\right)^2 + \frac{\lambda}{\Psi_G(p')}(a_{p'} - \tau_{p'})^2\right], \quad (21)$$

where $\Psi_G(p')$ is the proposed edge-aware weighting, $\tau_{p'}$ is the proposed edge-protect constraint. $\eta_{p'}$ is equal to $(1-\tau_{p'})$. It is the explicit constraint to control residual between input image and guidance image.

The optimal values of $a_{p'}$ and $b_{p'}$ are derived as follows:

$$a_{p'} = \frac{\eta(\mu_{G\odot X,\zeta}(p') - \mu_{G,\zeta}(p')\mu_{X,\zeta}(p')) + \frac{\lambda}{\Psi_G(p')}\tau_{p'}}{\eta\sigma_{G,\zeta}^2(p') + \frac{\lambda}{\Psi_G(p')}}, \quad (22)$$

$$b_{p'} = \mu_{X,\zeta}(p') - a_{p'}\mu_{G,\zeta}(p'), \quad (23)$$

For the final value of $\bar{a}_p, \bar{b}_p$, using the calculation format in (19), a weighted average strategy is proposed, which can be computed as follows:

$$\bar{a}_p = \frac{\sum_{p'\in\Omega_\zeta(p)}W_{p'}a_{p'}}{\sum_{p'\in\Omega_\zeta(p)}W_{p'}}; \bar{b}_p = \frac{\sum_{p'\in\Omega_\zeta(p)}W_{p'}b_{p'}}{\sum_{p'\in\Omega_\zeta(p)}W_{p'}}; \quad (24)$$

The filtering output $R$ is computed as follows:

$$R(p) = \bar{a}_p G(p) + \bar{b}_p, \quad (25)$$

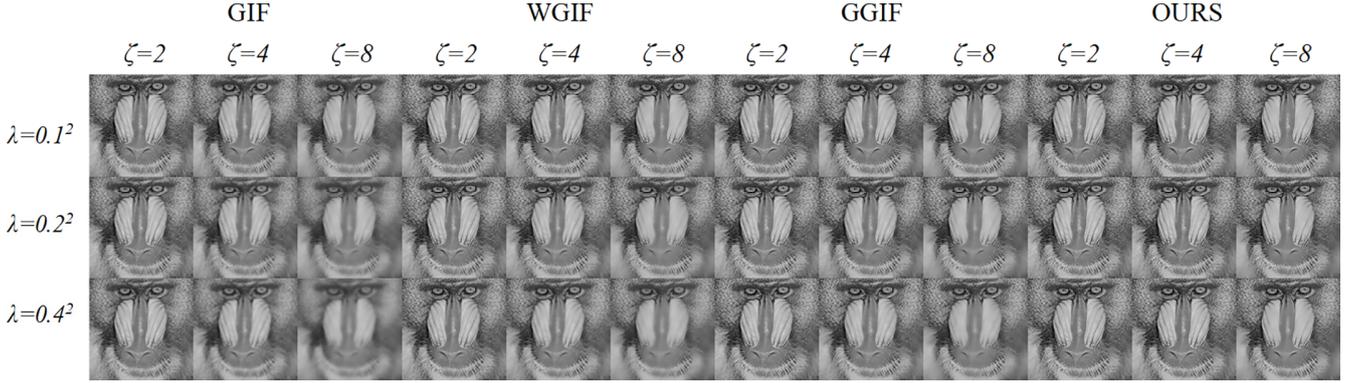

Fig. 3. Edge-preserving smoothing results by the GIF, the WGIF, the GGIF and Ours. The radius $\zeta$ and regularization parameter $\lambda$ are set to 2,4,8 and $0.1^2, 0.2^2, 0.4^2$, respectively.

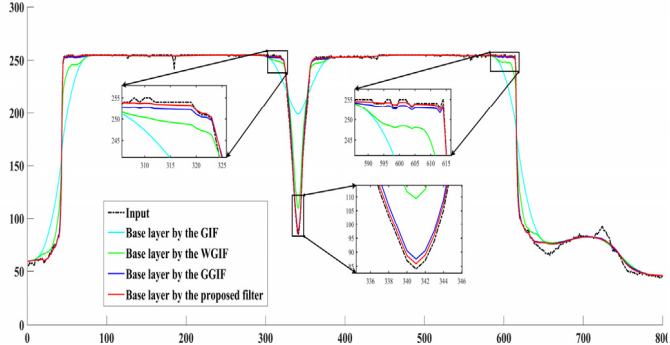

Fig. 4. 1D illustration of the GIF, the WGIF, the GGIF and Ours. $\zeta 1 = 16$, $\lambda = 1$ are set in all the algorithms. The input data is obtained from the row 608 of the red channel in Fig. 2 (a).

Table I
Quantitative assessment of results in Fig.3

|  | Methods | PSNR | | | SSIM | | |
|---|---|---|---|---|---|---|---|
|  |  | r=2 | r=4 | r=8 | r=2 | r=4 | r=8 |
| $0.1^2$ | GIF | 28.40 | 27.83 | 27.44 | 0.8272 | 0.8184 | 0.8358 |
|  | WGIF | 35.38 | 34.59 | 33.58 | 0.9117 | 0.9056 | 0.9126 |
|  | GGIF | 35.37 | 34.17 | 32.98 | 0.9115 | 0.8985 | 0.9015 |
|  | Ours | **38.73** | **38.28** | **37.86** | **0.9405** | **0.9402** | **0.9489** |
| $0.2^2$ | GIF | 24.04 | 23.15 | 22.50 | 0.6506 | 0.5987 | 0.5940 |
|  | WGIF | 31.34 | 30.40 | 29.24 | 0.8511 | 0.8331 | 0.8346 |
|  | GGIF | 32.69 | 31.23 | 29.72 | 0.8705 | 0.8446 | 0.8404 |
|  | Ours | **36.61** | **36.01** | **35.40** | **0.9174** | **0.9134** | **0.9216** |
| $0.4^2$ | GIF | 21.99 | 20.98 | 20.24 | 0.5086 | 0.4150 | 0.3807 |
|  | WGIF | 27.29 | 26.25 | 25.14 | 0.7555 | 0.7159 | 0.7050 |
|  | GGIF | 31.00 | 29.33 | 27.69 | 0.8380 | 0.8004 | 0.7883 |
|  | Ours | **34.46** | **33.80** | **33.07** | **0.8890** | **0.8799** | **0.8865** |

### E. Analysis of the proposed filter

For the convenience of analysis, it is assumed that images X and G are the same. Two cases are shown as follows:

*Case 1:* The pixel $p'$ is at an edge. The value of $\tau_{p'}$ is usually 1 and the value $\eta_{p'}$ is equal to 0. The value of $a_{p'}$ is computed as

$$a_{p'} = \frac{\frac{\lambda}{\Psi_G(p')}}{\frac{\lambda}{\Psi_G(p')}} = 1, \quad (26)$$

The value of $a_{p'}$ is 1 regardless of the value of $\lambda$, which means that sharp edges are preserved better than GIF, WGIF and GGIF even if the regularization parameter $\lambda$ is large.

*Case 2:* The pixel $p'$ is in a flat area. The value of $\tau_{p'}$ is usually 0 and the value $\eta_{p'}$ is equal to 1.
The value of $a_{p'}$ is computed as

$$a_{p'} = \frac{\sigma_{G,\zeta}^2(p')}{\sigma_{G,\zeta}^2(p') + \frac{\lambda}{\Psi_G(p')}}, \quad (27)$$

It can be seen that $a_{p'}$ is closer to 0 with the regularization parameter $\lambda$ is increased. This means the proposed filter can smooth the flat area even if regularization parameter $\lambda$ is large.

To verify the analysis above, some edge-preserving smoothing results are presented. As shown in Fig.3, the proposed method can preserve the sharp edges better than GIF, WGIF and GGIF. The PSNR and SSIM values of the results are reported in Table I. It can be seen that, with the same parameters, our method shows the highest scores among the related filters. In addition, as shown in Fig. 4, in comparison to the other three algorithms [16][17][18], sharp edges are preserved better by the proposed filter. From the three zoomed-in patches shown in the figure, the base layer processed by the proposed filter is much closer to the input data at the edges, so the proposed filter has a better performance of edge-preserving.

To further testing the effectiveness of our algorithm, as shown in Fig. 5, some real landscape photos were selected, and they all had clear edge areas. Larger radius ($\zeta 1 = 16$), larger regularization parameter ($\lambda = 0.1, 1, 10$), more algorithms [16][17][18][21][22][26] are selected into this experiment. As shown in Fig. 6, part of experiment results demonstrates that the comparison algorithms prone to produce halos near edges, where can be easily seen as the slightly red or white areas as pointed by the arrows. From the detail layer, blue areas can be obviously seen in the comparison algorithms and fewer edges are assigned to the detail layer by the proposed filter. when the regularization parameters are very large, the proposed filter can

J.Li et al.

filter can still preserve the sharp edges. Therefore, it can be intuitively judged that the proposed filter can better protect the edges. Furthermore, Table II shows the PSNR and SSIM values of the seven algorithms under different parameter settings in the whole experiment, among which the bold one is the best value, which shows that the proposed filter is better than the comparison filters.

When the guidance image is not identical with the input image, due to the large structural difference between the guidance image and the input image, this results in small or even negative covariances of the guidance image and the input iamge. It can be derived from (3) and (7) that the value of parameter $a_{p'}$ in GIF and WGIF will be very small, even negative, when the pixel is at an edge of the guidance image. In GGIF, as can be seen from (12), when the pixel is at an edge, even if the edge-protect constraint $\gamma_{p'}$ is equal to 1, due to the covariance in the numerator is not equal to the variance in the denominator, the value of parameter $a_{p'}$ is also not equal to 1. This seriously contradicts the idea of edge-preserving. However, benefited from the introduction of explicit residual constraint, as can be seen from (26), the value of parameter $a_{p'}$ is equal to 1, in other words, the proposed filter can effectively avoid edge-preserving degradation caused by inconsistent structure.

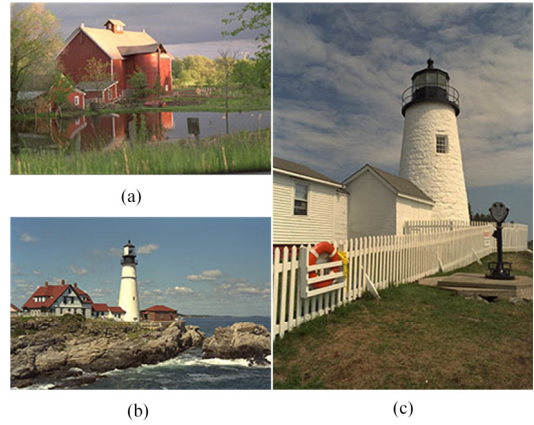

Fig. 5. Test images for edge-preserving smoothing[1]

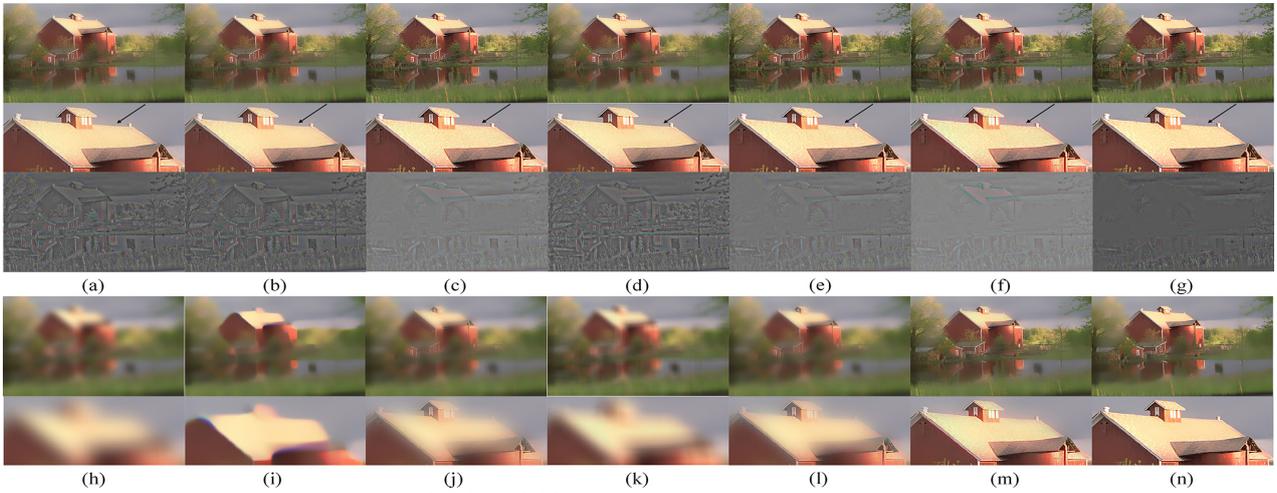

Fig. 6. Edge-preserving smoothing results. $\zeta=16$, $\lambda=0.01$ in (a)~(g) and $\lambda=1$ in (h)~(n). (a)~(g) and (h)~(n) are the results by the GIF, the WAGIF, the WGIF, the SKWGIF, the IWGIF, the GGIF and Ours, respectively. In (a)~(g), the images of each row are smooth results, zoom-in patches and detail layer of the results. In (h)~(n), the images of each row are smooth results and zoom-in patches, respectively.

Table II Quantitative assessment of results in Fig.5

| Methods | Images | PSNR | | | | | | SSIM | | | | | |
|---|---|---|---|---|---|---|---|---|---|---|---|---|---|
| | | $\lambda=0.05^2$ | $\lambda=0.1^2$ | $\lambda=0.1$ | $\lambda=1$ | $\lambda=10$ | average | $\lambda=0.05^2$ | $\lambda=0.1^2$ | $\lambda=0.1$ | $\lambda=1$ | $\lambda=10$ | average |
| GIF[16] | (a) | 34.34 | 28.42 | 23.44 | 22.22 | 22.07 | 26.10 | 0.9752 | 0.9236 | 0.8129 | 0.7771 | 0.7726 | 0.8523 |
| | (b) | 36.78 | 29.59 | 21.86 | 19.75 | 19.46 | 25.49 | 0.9804 | 0.9374 | 0.7688 | 0.6797 | 0.6666 | 0.8066 |
| | (c) | 35.19 | 29.42 | 22.32 | 20.08 | 19.78 | 25.36 | 0.9729 | 0.9335 | 0.8260 | 0.7636 | 0.7540 | 0.8500 |
| WAGIF[22] | (a) | 34.35 | 28.44 | 23.54 | 22.52 | 22.41 | 26.25 | 0.9753 | 0.9236 | 0.8208 | 0.7946 | 0.7919 | 0.8612 |
| | (b) | 36.81 | 29.64 | 21.77 | 19.48 | 19.25 | 25.39 | 0.9804 | 0.9379 | 0.7636 | 0.6898 | 0.6818 | 0.8107 |
| | (c) | 35.21 | 29.46 | 22.30 | 20.06 | 19.81 | 25.37 | 0.9729 | 0.9337 | 0.8235 | 0.7693 | 0.7627 | 0.8524 |
| WGIF[17] | (a) | 40.97 | 35.96 | 29.42 | 25.29 | 23.12 | 30.95 | 0.9902 | 0.9786 | 0.9311 | 0.8542 | 0.7973 | 0.9103 |
| | (b) | **42.56** | 39.40 | 33.38 | 26.91 | 22.03 | 32.86 | 0.9917 | 0.9845 | 0.9586 | 0.8876 | 0.7651 | 0.9175 |
| | (c) | 39.47 | 36.27 | 30.87 | 26.26 | 22.16 | 31.00 | **0.9816** | 0.9682 | 0.9337 | 0.8841 | 0.8163 | 0.9168 |
| SKWGIF[21] | (a) | 34.83 | 29.09 | 24.05 | 22.64 | 22.44 | 26.61 | 0.9752 | 0.9274 | 0.8241 | 0.7872 | 0.7821 | 0.8592 |
| | (b) | 37.39 | 30.42 | 22.60 | 20.25 | 19.91 | 26.11 | 0.9803 | 0.9394 | 0.7820 | 0.6910 | 0.6767 | 0.8139 |
| | (c) | 35.63 | 30.12 | 22.89 | 20.34 | 19.97 | 25.79 | 0.9724 | 0.9361 | 0.8373 | 0.7707 | 0.7595 | 0.8552 |
| IWGIF[26] | (a) | 41.22 | 36.22 | 29.70 | 25.62 | 23.41 | 31.23 | 0.9904 | 0.9793 | 0.9341 | 0.8604 | 0.8050 | 0.9138 |
| | (b) | 42.86 | 39.80 | 33.99 | 27.59 | 22.61 | 33.37 | 0.9919 | 0.9849 | 0.9605 | 0.8951 | 0.7779 | 0.9221 |
| | (c) | 39.50 | 36.36 | 31.19 | 26.82 | 22.62 | 31.30 | **0.9816** | 0.9684 | 0.9355 | 0.8896 | 0.8241 | 0.9198 |
| GGIF[18] | (a) | 39.61 | 35.53 | 31.06 | 29.04 | 28.38 | 32.72 | 0.9892 | 0.9778 | 0.9497 | 0.9274 | **0.9181** | 0.9524 |
| | (b) | 41.47 | 39.30 | 35.07 | 30.52 | 28.33 | 34.94 | **0.9922** | **0.9867** | 0.9684 | 0.9354 | 0.9128 | 0.9591 |
| | (c) | 39.16 | 35.63 | 30.77 | 28.11 | 27.26 | 32.19 | 0.9805 | 0.9660 | 0.9325 | 0.9092 | 0.9022 | 0.9381 |
| Ours | (a) | **42.95** | **39.58** | **34.56** | **30.56** | **28.66** | **35.26** | 0.9910 | 0.9863 | 0.9697 | 0.9372 | 0.9101 | **0.9589** |
| | (b) | 42.31 | **40.78** | **38.63** | **36.12** | **32.89** | **38.15** | 0.9891 | 0.9845 | **0.9761** | **0.9640** | **0.9430** | **0.9713** |
| | (c) | **39.82** | **37.94** | **34.83** | **31.70** | **30.05** | **34.87** | 0.9809 | **0.9728** | **0.9580** | **0.9361** | **0.9189** | **0.9533** |

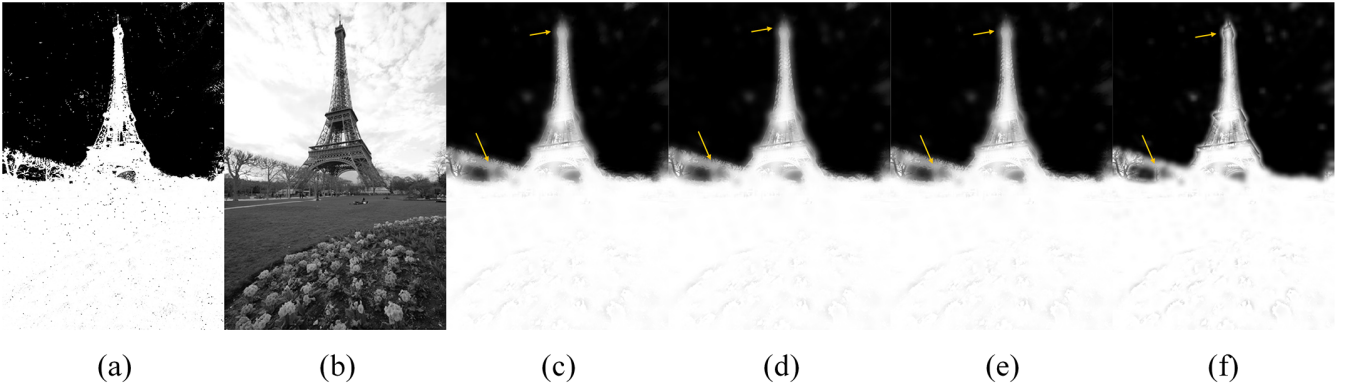

(a)        (b)        (c)        (d)        (e)        (f)

Fig. 7. Edge-preserving smoothing results for inconsistent structure. $\zeta=4$, and $\lambda=1/1024$. (a) input image (b)guidance image (c) GIF (d) WGIF (e) GGIF (f) Ours.

To verify the analysis above, as shown in Fig.7, more edges are preserved by the proposed filter, which are pointed by the yellow arrows.

---

**Algorithm 1**: enhanced edge-perceptual guided image filter

**Input:** input image $X$, guidance image $G$, set $\zeta$, $\lambda$ and $c$.
**Output:** output image R

1: $\text{mean}_T = F_{\text{mean}}(T)$ with $T=X,G$
   $\text{corr}_S = F_{\text{mean}}(S)$ with $S = X.*X,\ G.*G,\ G.*X$

2: $\text{var}_X = \text{corr}_X - \text{mean}_X.*\text{mean}_X$
   $\text{var}_G = \text{corr}_G - \text{mean}_G.*\text{mean}_G$
   $\text{cov}_{G.*X} = \text{corr}_{G.*X} - \text{mean}_G.*\text{mean}_X$

3: Compute $\Psi_G(p')$ using (15)
   Compute $\tau_{p'}$ using (16) and (17)

4: Compute $a_{p'}$ and $b_{p'}$ using (22) and (23)

5: Compute $W_{p'}$ using (19)

6: $N = (2\zeta+1)^2$
   $W_{\text{sum}} = \text{boxfilter}(W_{p'}, N)$

   $a^*_{p'} = \text{boxfilter}(W_{p'}.*a_{p'}, N)$

   $b^*_{p'} = \text{boxfilter}(W_{p'}.*b_{p'}, N)$

7: $R = (a^*_{p'}.*G + b^*_{p'})./W_{\text{sum}}$

/* $F_{\text{mean}}$ refers to boxfilter(data, N)./N */

---

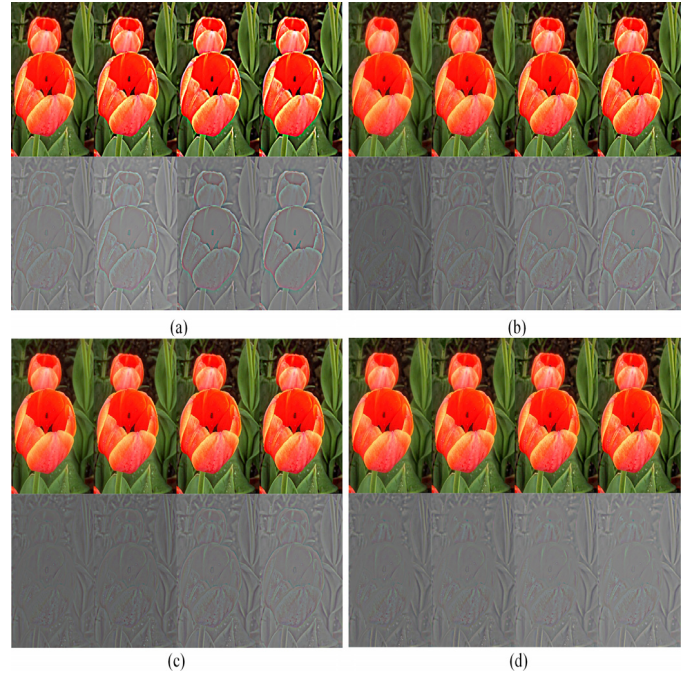

(a)        (b)

(c)        (d)

Fig. 8. Comparison of image enhancement results. The images of each row are enhanced results and detail layers in (a)~(c). From left to right, $\lambda = 0.05^2, 0.1^2, 0.2^2, 0.1$ in each set of results. (a) GIF (b) WGIF (c) GGIF (d) Ours.

## IV. APPLICATIONS

In this section, the proposed gradient domain content adaptive guided image filter is adopted to study many applications, such as single image detail enhancement, multi-scale exposure fusion and hyper spectral image classification.

### A. Single image detail enhancement

Single image detail enhancement is a typical application in image processing, which can comprehensively evaluate the performance of the filter in surpressing halos and gradient reversal artifacts. A base layer image can be obtained after edge-preserving smoothing, and a detail layer image can be obtained by subtracting the base layer image from the input image. The enhanced image can be obtained by adding five times detail layer image to the input image.

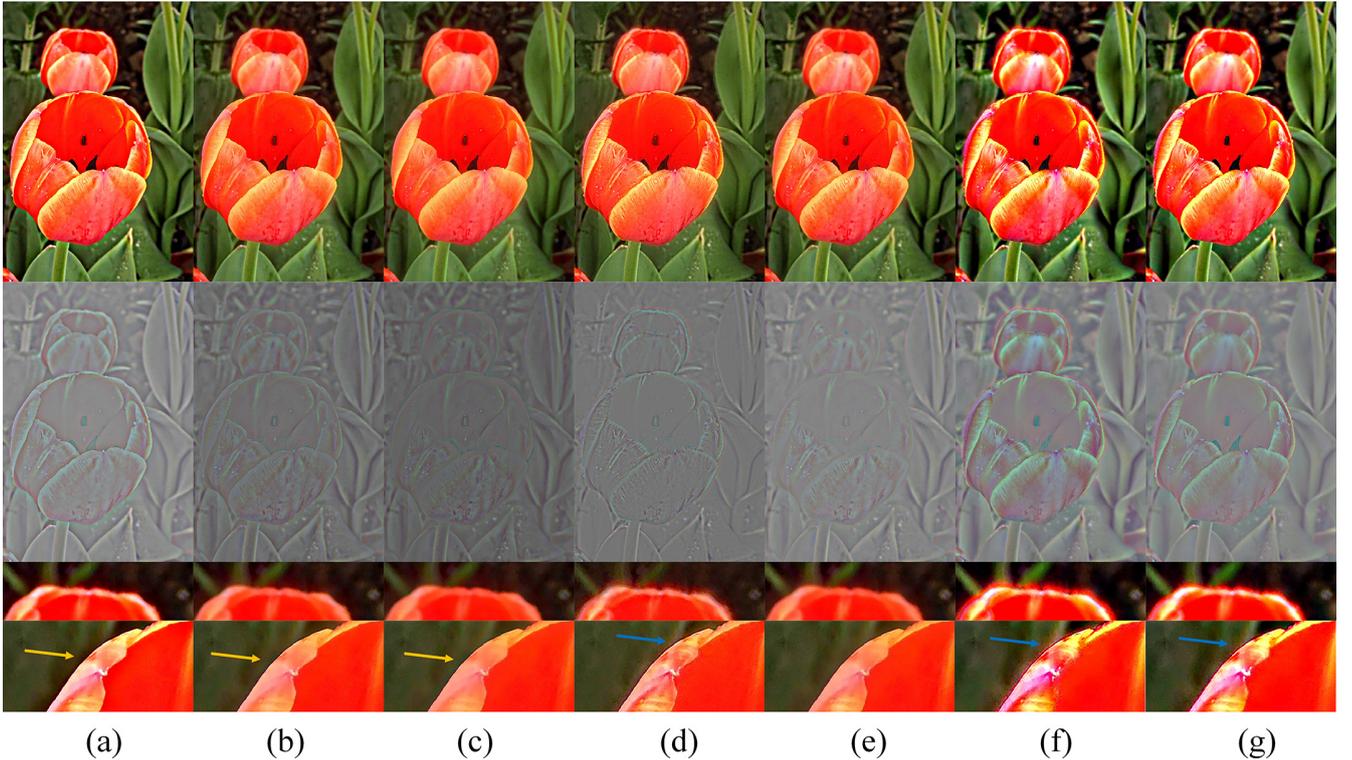

(a)     (b)     (c)     (d)     (e)     (f)     (g)

Fig. 9. Comparison of detail enhancement. The images of each row are enhanced results, detail layers, two sets of zoomed-in patches from enhanced results, respectively. $\zeta = 16$ in the GIF, WGIF, GGIF, AnisGF, the proposed filter. $\lambda = 0.1^2$ in the GIF, WGIF, GGIF, AnisGF, and $\lambda = 0.1$ in the proposed filter. $\lambda = 0.1^2$, $\sigma_1 = 0.5$, $\sigma_2 = 1$ in the ROG. $\lambda = 1$, $\alpha = 0.2$, $r_s = 16$ in the GSF. (a) GIF (b) WGIF (c) GGIF (d) AnisGF (d) Ours (f) ROG (g) GSF.

Table III
Quantitative assessment of Fig.8

|  | BIQI | | | | NIQE | | | |
|---|---|---|---|---|---|---|---|---|
| $\lambda$ | $0.05^2$ | $0.1^2$ | $0.2^2$ | $0.1$ | $0.05^2$ | $0.1^2$ | $0.2^2$ | $0.1$ |
| GIF | 34.43 | 33.04 | 33.10 | 31.72 | 5.03 | 4.95 | 4.95 | 4.97 |
| WGIF | 35.36 | 34.77 | 33.20 | 32.38 | 5.16 | 5.02 | 4.97 | 4.97 |
| GGIF | 35.69 | 35.95 | 34.81 | 34.06 | 5.20 | 5.05 | 5.00 | 4.95 |
| Ours | **38.49** | **39.63** | **39.49** | **38.79** | 5.04 | 4.97 | **4.89** | **4.87** |

Three influential filters [16][17][18] are selected into the experiment. As shown in Fig.8, with the increase of $\lambda$, there will be more details in the detail layer. However, it will introduce more annoying halos at the same time. Compared with the other three filters, the proposed filter can introduce less halos at the edges which can be easily seen in the detail layer images of each set of images. It is worth mentioning that the proposed filter assigns less edges into detail layer image compared with the other three filters, which will be helpful for image detail enhancement by using the proposed filter when $\lambda$ is a large number.

In addition to using subjective sensory evaluation, objective parameter evaluation is applied to characterize the superiority of the proposed filter. There are two blind image evaluation indexes are selected, which are Blind Image Quality Index (BIQI) in [27] and Natural Image Quality Evaluator (NIQE) in [28]. With the metric BIQI, a higher value represents better performance. Moreover, with the metric NIQE, a lower value represents better performance. From the Table III, the BIQI scores at four different setting of the proposed filter is higher than the GIF, the WGIF and the GGIF, which prove the higher image quality processed by the proposed filter. For the index of NIQE, when the $\lambda$ is large, the scores of the proposed filter are lower than other three filters. This shows that the enhanced image obtained by the proposed filter is closer to the real scene.

To verify the efficiency of the proposed filter, the GIF in [16], the WGIF in [17], the GGIF in [18], the AnisGF in [20], the ROG in [29], the GSF in [30] are selected for the image enhancement. From the detail enhanced image presented in the first row of the Fig.9, the proposed filter and the AnisGF have better detail-enhanced performance and remarkable visual effect. From the detail layer image presented in the second row of the Fig.9, fewer strong edges penetrate to the detail layers by the proposed filter, and less halos happened near edges in the detail layers at the same time. From the two sets of the zoom-in patches, it can be seen that (a)-(c) have halos (little darker areas near edges) to some degree as pointed by the yellow arrows. And (d), (f), (g) have gradient reversal artifacts near edges as pointed by the blue arrows. It is worth mentioning that the $\lambda$ in the proposed filter is much larger than the GIF, the WGIF, the GGIF and the AnisGF. This means that a large $\lambda$ can be selected without worrying the annoy artifacts.

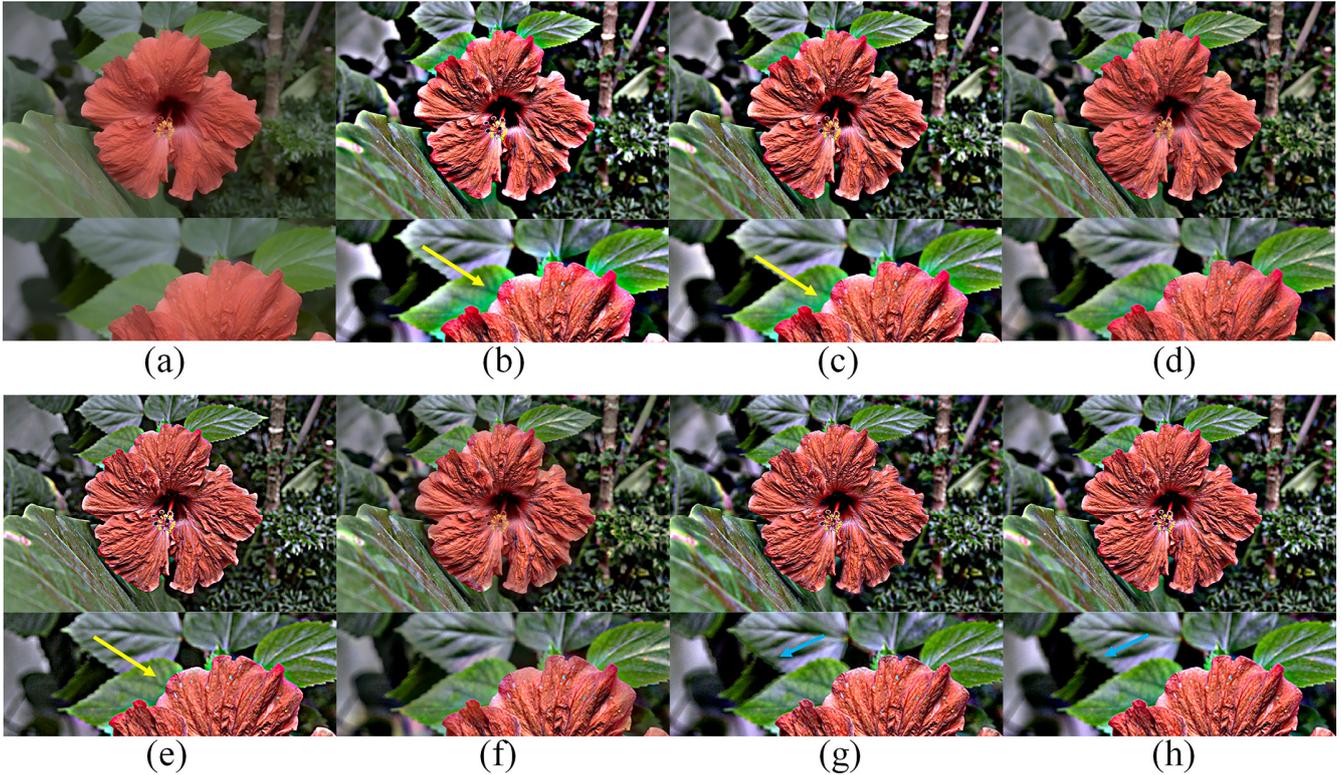

Fig. 10. Comparison of detail enhancement. The images of each row are enhanced results, detail layers, two sets of zoomed-in patches from enhanced results, respectively. $\zeta = 16$ in the GIF, WGIF, GGIF, AnisGF, the proposed filter. $\lambda = 1$ in the GIF, WGIF, GGIF, AnisGF, and the proposed filter. $\lambda = 0.1^2$, $\sigma_1 = 0.5$, $\sigma_2 = 1$ in the ROG. $\lambda = 1$, $\alpha = 0.2$, $r_s = 16$ in the GSF. (a) input (b) GIF (c) WGIF (d) GGIF (e) AnisGF (f) Ours (g) ROG (h) GSF.

Table IV
Quantitative assessment of Fig.9 and Fig.10

|  | BIQI | | | NIQE | | |
| --- | --- | --- | --- | --- | --- | --- |
|  | Fig.9 | Fig.10 | Average | Fig.9 | Fig.10 | Average |
| GIF | 33.04 | 29.74 | 31.39 | 4.95 | 3.06 | 4.01 |
| WGIF | 34.77 | 32.27 | 33.52 | 5.02 | 3.02 | 4.02 |
| GGIF | 35.95 | 39.38 | 37.67 | 5.05 | 3.06 | 4.06 |
| AnisGF | 31.20 | 26.92 | 29.06 | 4.52 | 3.06 | 3.79 |
| Ours | 38.79 | 37.80 | **38.30** | 4.87 | 2.84 | 3.86 |
| ROG | 38.43 | 24.89 | 31.66 | 4.53 | 2.73 | **3.63** |
| GSF | 34.79 | 26.32 | 30.56 | 4.55 | 2.80 | 3.68 |

As shown in Fig.10, a large $\lambda$ is setting to test the detail enhancement performance of the different filters. It can be seen from (b), (c) and (e) that halos are prone to produce near the edges of the petal. In (g) and (h), leaf edges are prone to over-enhanced and thus gradient reversal happened. The proposed filter can enhance the image while suppressing artifacts.

Similarly, two indicators are used to measure the enhancement effect of the image in Fig.9 and Fig.10. However, it is worth mentioning that the $\lambda$ is not the best setting of the guided filter and its improved version, the parameters in (g) and (h) are commonly used in the paper[29][30]. From Table IV, the BIQI scores of the proposed filter are almost the highest of the seven filters, and the average score is the highest. This shows that the proposed filter performs better and has higher stability in different scenarios. The NIQE scores of the proposed filter are lower than the GIF, the WGIF, the GGIF and closer to the methods based on global optimization.

### B. Multi-scale exposure fusion

Multi-scale exposure fusion is commonly applied in the High Dynamic Range (HDR) imaging technique, which can effectively improve the image quality. In the exposure fusion algorithms based on image pyramids, edge-preserving filters are often used to smooth the weight maps while preserving more edge information[4][33][35]. To verify the effectiveness of the proposed filter in this application, five algorithms[4][32][33][34][35] were selected for comparison. Similar to the algorithm [4], under the basic framework of the algorithm [32], the proposed filter is applied to smooth the weight maps by using the brightness image of the source image as a guidance image. 17 sets of multi-exposure image sequences were tested, and commonly used MEF-SSIM in [31] was selected as the evaluation standard. Its value is between 0~1, and higher value

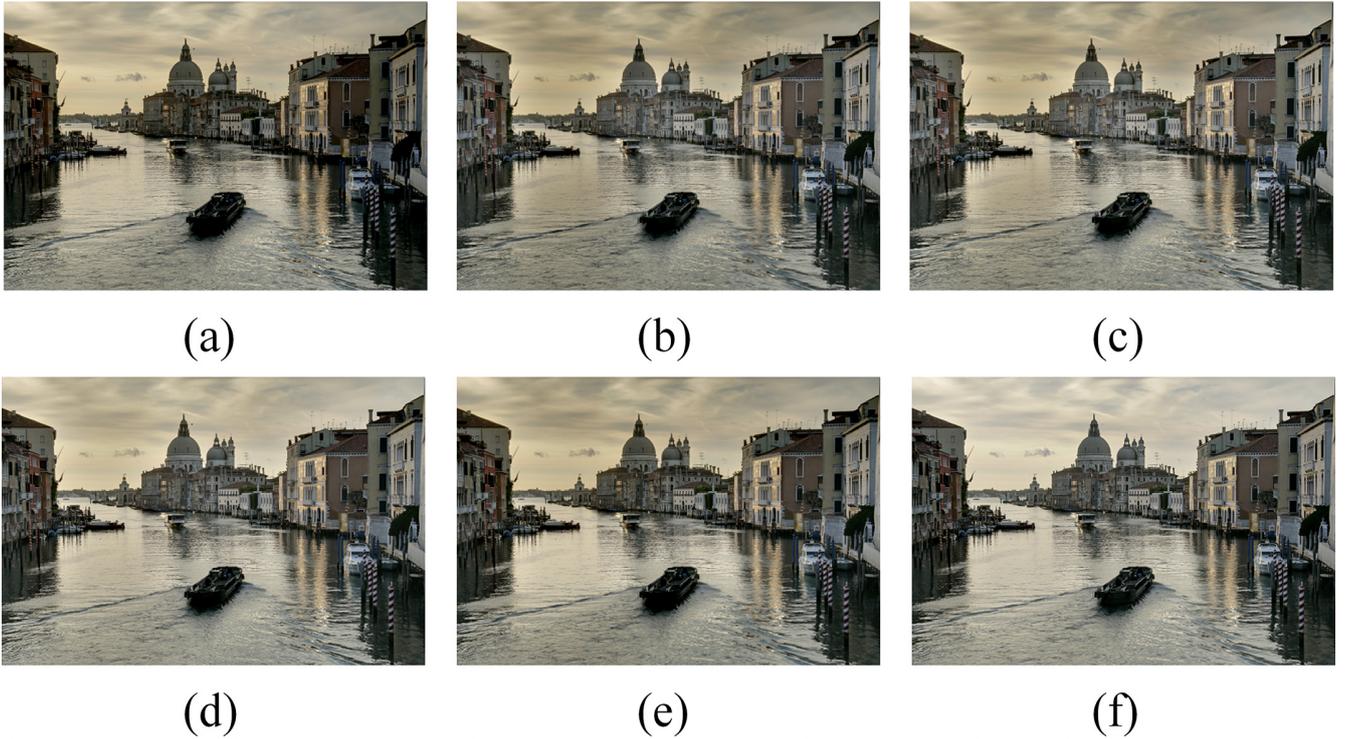

Fig. 11. Comparison of multi-scale exposure fusion. (a) [32] (b) [33] (c) [4] (d) [34] (e) [35] (f) Ours (*β*=1/50)

Table V
Quantitative assessment of exposure fusion

|  | [32] | [33] | [4] | [34] | [35] | Ours |
|---|---|---|---|---|---|---|
| Cave | 0.9748 | 0.9729 | 0.9785 | 0.9775 | **0.9831** | 0.9825 |
| Tower | 0.9857 | 0.9841 | 0.9862 | 0.9877 | **0.9882** | 0.9878 |
| House | 0.9643 | 0.9601 | 0.9613 | 0.9600 | **0.9650** | 0.9638 |
| Balloons | **0.9694** | 0.9469 | 0.9514 | 0.9572 | 0.9660 | 0.9686 |
| Venice | 0.9661 | 0.9509 | 0.9521 | 0.9631 | 0.9656 | **0.9702** |
| Office | 0.9845 | 0.9808 | 0.9839 | 0.9856 | **0.9899** | 0.9858 |
| Lighthouse | **0.9802** | 0.9702 | 0.9703 | 0.9742 | 0.9759 | 0.9787 |
| Landscape | **0.9762** | 0.9462 | 0.9474 | 0.9743 | 0.9596 | 0.9685 |
| Kluki | **0.9801** | 0.9699 | 0.9711 | 0.9764 | 0.9797 | 0.9796 |
| farmhouse | 0.9810 | 0.9821 | 0.9816 | **0.9831** | 0.9798 | 0.9810 |
| garden | 0.9889 | 0.9817 | 0.9827 | 0.9854 | 0.9863 | **0.9905** |
| Belgium | 0.9709 | 0.9675 | 0.9680 | 0.9729 | 0.9719 | **0.9735** |
| Cadik | **0.9690** | 0.9457 | 0.9453 | 0.9424 | 0.9583 | 0.9615 |
| Candle | **0.9712** | 0.9453 | 0.9503 | 0.9480 | 0.9675 | 0.9648 |
| Madison | 0.9773 | 0.9674 | 0.9692 | 0.9724 | 0.9780 | **0.9798** |
| Memorial | 0.9670 | 0.9690 | 0.9701 | 0.9696 | 0.9689 | **0.9702** |
| Lamp | 0.9476 | 0.9346 | 0.9389 | 0.9386 | **0.9521** | 0.9516 |
| **Average** | 0.9738 | 0.9633 | 0.9652 | 0.9687 | 0.9727 | **0.9740** |

represents higher image quality.

As shown in Fig.11, the buildings are dark in (a), and halos are prone to produce at the junction of the sky and buildings in (b)-(d). Compared with (e), the proposed algorithm is brighter, and the junction of the sky and buildings is more natural.

It can be found from Table V that the proposed algorithm has the highest average value in the 17 sets of multi-exposure image sequences, which means that the proposed filter can perform better in this task.

### C. Hyper spectral image classification

A hyper spectral image has hundreds of spectral bands for each pixel, which means that abundant spectral information can be obtained from the image. An edge-preserving filtering based spectral-spatial classification framework is proposed in [37], which broadened the application of this kind of filter. The framework first utilizes SVM tool to train the initial probability

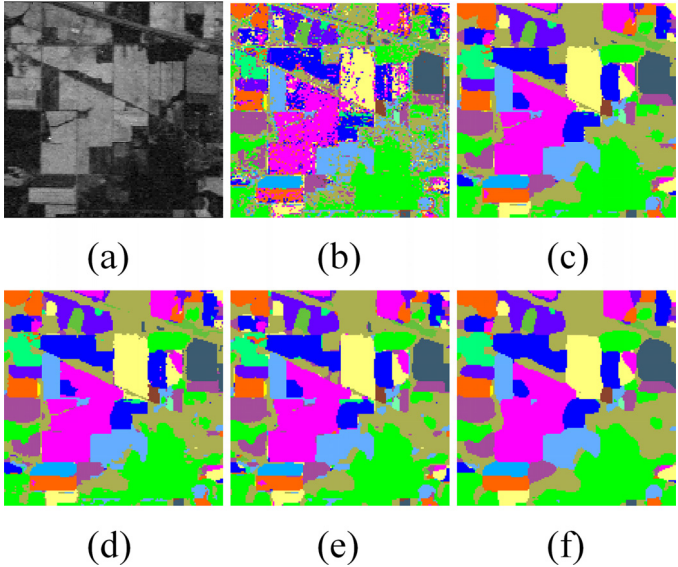

Fig. 12. Comparison of classification results (Indian Pines). $\sigma_s$=3, $\sigma_r$=0.1 in the bilateral filter and the NC filter. $\zeta$=3, and $\lambda$=0.001 in the guided filter. $\zeta$=3, and $\lambda$=0.1 in the proposed filter. (a) remote-sensing image (b) initial probability maps (c) bilateral filter (d) guided filter (e) NC filter (f) Ours

maps. And the initial probability maps can be smoothed by the edge-preserving filters, with the component of the hyper spectral image as the guidance image. Finally, the classification result is determined by the filtered image. It can be seen that the quality of the filtered image directly determines the accuracy of the classification results.

The proposed filter can be easily applied in this framework. As shown in Fig.12, compared with the bilateral filter [13] based in (c), the guided filter [16] based in (d) and the NC filter [36] based in (e), the proposed filter based in (f) can better smooth the probability maps, and the classification results is more likely the original spectral image. Moreover, the data in Table VI shows that the classification results based by the proposed filter is more accurate than other three filters.

## V. CONCLUSION

In this paper, we systematically analyzed the solution to the problem that guided image filter is prone to produce halos at the edges, and proposed a new constraint to improve the edge-preserving ability. Besides, a new residual constraint was proposed to improve the edge-preserving ability in the case of the inconsistent structure between the input image and the guidance image. In these three typical applications, the proposed filter also exhibits good edge-preserving ability. How to design the algorithm framework more reasonably so that the proposed filter can be better applied to these fields. It will be studied in our future research.

ACKNOWLEDGEMENTS

Table VI
Accuracy of the classification results by different filters

| Class | BF | GF | NC | Ours |
|---|---|---|---|---|
| Alfalfa | 1.0000 | 1.0000 | 1.0000 | 1.0000 |
| Corn-N | 0.9768 | 0.9772 | 0.9747 | 0.9770 |
| Corn-M | 0.8954 | 0.9056 | 0.8936 | 0.9065 |
| Corn | 0.6484 | 0.6043 | 0.6185 | 0.6844 |
| Grass-M | 0.9419 | 0.9447 | 0.9554 | 0.9828 |
| Grass-T | 0.9939 | 0.9985 | 1.0000 | 0.9939 |
| Grass-P-M | 1.0000 | 1.0000 | 1.0000 | 1.0000 |
| Hay-W | 1.0000 | 1.0000 | 1.0000 | 1.0000 |
| Oats | 1.0000 | 1.0000 | 1.0000 | 1.0000 |
| Soybean-N | 0.9264 | 0.9296 | 0.9294 | 0.9288 |
| Soybean-M | 0.9693 | 0.9727 | 0.9768 | 0.9793 |
| Soybean-C | 0.9268 | 0.9131 | 0.8867 | 0.9265 |
| Wheat | 1.0000 | 1.0000 | 1.0000 | 1.0000 |
| Woods | 0.9965 | 0.9947 | 0.9946 | 0.9825 |
| Bulidings | 0.8035 | 0.8076 | 0.7857 | 0.7838 |
| Stone-S-T | 0.9216 | 0.9400 | 0.9400 | 0.9583 |
| OA | 0.9445 | 0.9466 | 0.9469 | **0.9512** |
| AA | 0.9347 | 0.9368 | 0.9375 | **0.9440** |
| Kappa | 0.9364 | 0.9387 | 0.9391 | **0.9440** |